\documentclass[]{style/ceurart}
\usepackage{listings}
\usepackage{tikz}
\usetikzlibrary{shapes.geometric, arrows.meta, positioning, calc}

\begin{document}

\copyrightyear{2025}
\copyrightclause{Copyright for this paper by its authors.
  Use permitted under Creative Commons License Attribution 4.0
  International (CC BY 4.0).}

\conference{CLEF 2026: Conference and Labs of the Evaluation Forum, September 21--24, 2026, Jena, Germany}

\title{DS@GT ARC at ImageCLEFmedical 2026: Architectural Diversity for Concept Detection and Foundation-Model Scaling for Caption Prediction in  Medical Image Analysis}

\author[1]{Bowen Wang}[
    orcid=0009-0004-8132-1769,
    email=bwang685@gatech.edu,
]
\cormark[1]

\author[1]{Youwen Zhang}[
    orcid=0000-0002-0957-4276,
    email=yzhang3925@gatech.edu,    
]
\cormark[1]

\author[1]{Ritesh Mehta}[
    orcid=0009-0000-2786-1626,
    email=rmehta307@gatech.edu,
]
\cormark[1]

\address[1]{Georgia Institute of Technology, North Ave NW, Atlanta, GA 30332}
\cortext[1]{Corresponding author.}

\begin{abstract}
We describe the DS@GT submissions to the ImageCLEFmedical Caption 2026 challenge, which continues a long-running benchmark on the ROCOv2 dataset with two tracks: Concept Detection (Task~1), assigning UMLS Concept Unique Identifiers (CUIs) to radiology images, and Caption Prediction (Task~2), generating natural-language captions.
For Task~1, our primary submission was a three-way late-fusion ensemble of ConvNeXt-V2, BiomedCLIP ViT-B/16, and DenseNet-169 with a regularized ``Honest Threshold Tuning'' procedure designed to avoid validation overfitting on rare concepts; this submission ranked first on the official submission with a primary $F_1$ of $0.5790$ and a secondary $F_1$ of $0.9657$.
In parallel, we submitted a training-free KNN retrieval pipeline over frozen BiomedCLIP embeddings, which reached a primary $F_1$ of $0.5780$ and a secondary $F_1$ of $0.9599$---essentially matching the fine-tuned ensemble on the primary track at a fraction of the cost.
For Task 2, our submissions included a fine-tuned Gemma-3 27B model (overall $0.3571$, ranking third in the official submission), a fully fine-tuned BLIP pipeline with custom Vizwins merging ($0.3564$), and a zero-shot MedGemma-4B run with a PubMed-style prompt ($0.3186$), spanning a wide range of model scales and training costs. Code: https://github.com/dsgt-arc/imageclef-caption-2026.
\end{abstract}


\maketitle

\section{Introduction}

The ImageCLEFmedical Caption 2026 task challenges systems to automatically interpret radiology images through two primary tracks: Concept Detection (Task~1), which assigns relevant Unified Medical Language System (UMLS)~\cite{bodenreider2004umls} Concept Unique Identifiers (CUIs), and Caption Prediction (Task~2), which generates clinically meaningful descriptions in natural-language. Both tasks utilize an updated and extended version of the ROCOv2~\cite{ruckert2024rocov2}, which presents significant machine learning challenges, including extreme long-tailed distributions in medical concepts and the inherent trade-off between clinical factuality and natural language fluency in Large Vision-Language Models (VLMs). 
We refer the reader to the ImageCLEF 2026 lab overview~\cite{ImageCLEF2026} and the ImageCLEFmedical 2026 task overview~\cite{ImageCLEFmedicalCaptionOverview2026} for the full task description and shared evaluation setup.

These working notes detail our team's methodology and top-performing submissions. On the official ImageCLEFmedical 2026 leaderboard our submissions are reported under the username \texttt{dsgt\_caption}. For Task~1, we developed a highly robust pipeline that achieved the first place ranking on the official test set. Rather than relying on a standard global threshold, our success was driven by a novel ``Honest Threshold Tuning'' strategy applied to a three-way ensemble of vision encoders (ConvNeXt-V2~\cite{woo2023convnextv2}, BiomedCLIP~\cite{zhang2023biomedclip}, and DenseNet-169~\cite{huang2017densenet}). By pruning the label space and applying strict mathematical regularization (long-tail fallbacks and threshold blending), we successfully mitigated the validation overfitting commonly seen in long-tail medical distributions.

For Task~2, our strongest submission was a QLoRA fine-tuned Gemma-3 27B model decoded with 3-beam search.
Alongside this we explored two contrasting lines: an architectural ablation of \emph{concept anchoring}---injecting Task~1 predictions as structured prompts into smaller VLMs (InstructBLIP~\cite{dai2023instructblip}, LLaVA-Med~\cite{li2023llavamed}, BLIP~\cite{li2022blip})---with an expanded generation budget and a custom caption-merging algorithm, and a zero-shot MedGemma-4B run with a PubMed-style legend prompt that requires no fine-tuning at all.
Together these submissions span roughly a 100$\times$ range in model size, and the contrast between them motivates the central observation of our Discussion: for medical caption generation, foundation-model scale and pretraining-distribution match appear to be the dominant levers, while concept anchoring is a complementary strategy for smaller, resource-constrained models.

\section{Related Work}

The ImageCLEFmedical Caption task was first proposed in 2016~\cite{herrera2016imageclef}. In 2017 and 2018, it comprised two main subtasks: concept detection and caption prediction~\cite{eickhoff2017imageclef,garcia2018imageclef}. While the evaluation focus temporarily shifted exclusively to concept detection from 2019 to 2020~\cite{pelka2019imageclef,pelka2020imageclef}, both subtasks have run in parallel again since 2021~\cite{pelka2021imageclef}. Recent editions have continued to refine the benchmark with higher-quality, manually annotated data and updated evaluation metrics for caption prediction~\cite{rueckert2023imageclef}.

For concept detection, top-performing systems have consistently relied on ensembles of Convolutional Neural Networks (CNNs). In 2025, the winning AUEB NLP Group~\cite{auebnlpcaption2025} combined EfficientNet-B0~\cite{tan2019efficientnet}, DenseNet-121~\cite{huang2017densenet}, and ConvNeXt-Tiny~\cite{liu2022convnext} with per-label threshold optimization and dual-threshold aggregation. Our approach builds directly on their threshold-tuning strategy, adding a regularization layer to prevent overfitting on rare concepts.

For caption prediction, the 2025 task saw a clear shift toward fine-tuned VLMs. The winning UMUTeam~\cite{umuteam2025} fine-tuned a BLIP~\cite{li2022blip} model with full-weight adaptation, selecting the best checkpoint based on the relevance average. DS4DH~\cite{ds4dh2025} explored retrieval-augmented generation (RAG) with InstructBLIP~\cite{dai2023instructblip}, finding that cluster-based RAG using MedCPT~\cite{jin2023medcpt} embeddings outperformed standard nearest-neighbor retrieval, but that fine-tuned InstructBLIP without retrieval remained the strongest single system. The 2025 results demonstrated that models with direct fine-tuning on medical data and minimal architectural complexity often outperformed more elaborate multi-stage pipelines, a finding that informed our own approach.

Domain-adapted vision encoders have proven valuable for medical multi-label classification.
BiomedCLIP~\cite{zhang2023biomedclip}, pretrained on fifteen million biomedical image-text pairs from PubMed Central, provides strong visual priors for modality identification.
It is usually used as a frozen feature extractor inside an ensemble, but its embedding space already carries a lot of useful medical structure on its own.
Also, work on clustering DINO features for anomaly detection~\cite{Schulthess2025AnomalyDB} has shown that single dimensions of big pretrained vision encoders can be surprisingly class-discriminative without any extra training.
The Platonic Representation Hypothesis~\cite{Huh2024ThePR} makes a broader claim in the same direction: models trained on enough data tend to learn similar internal geometries, which means simple linear probes can go a long way.
We build on this idea when characterizing the ROCOv2 [2] dataset, as shown in Section 3.1.
Asymmetric Loss (ASL)~\cite{benbaruch2021asymmetricloss} has emerged as a standard technique for handling label imbalance in multi-label settings, down-weighting easy negatives while preserving gradient signal on rare positives.

For caption generation, concept anchoring, injecting structured medical concept predictions into the prompt, has been explored as a way to improve clinical factuality.
The AUEB NLP Group~\cite{auebnlpcaption2025} used a synthesizer and LM-Fuser approach to incorporate concept information into InstructBLIP outputs.
Our work extends this direction with a systematic cross-architecture comparison of how different model families respond to structured concept conditioning.

Another way to do captioning, different from training a full VLM end-to-end, is to take a vision encoder and a language model that were trained separately and just connect them with a small learned mapping.
Generating Text from Images in a Smooth Representation Space~\cite{Spinks2018GeneratingTF} does this by projecting image embeddings into the input space of a language model, and Harnessing the Universal Geometry of Embeddings~\cite{Jha2025HarnessingTU} shows that the mapping can be very light when the two embedding spaces already line up well.

General-purpose VLMs surveyed in~\cite{Zhang2023VisionLanguageMF} are strong on natural images, but they have some pretty clear weak spots on radiology.
A recent paper even shows that VLMs often can't tell left from right in medical images~\cite{Wolf2025YourOL}, which is a real problem when you try to use them zero-shot on ROCOv2 captions.
One way people have tried to fix this is by feeding clinical concepts into the language model as extra context~\cite{Sha2025ContrastiveKL}, which is similar in spirit to the concept-anchoring used in this paper.
Our MedGemma fine-tuning experiments, presented in Section 3.4, look at how parameter-efficient adaptation and joint concept/caption training shift this trade-off on the ImageCLEFmedical 2026 data.

\section{Methodology}

\subsection{Dataset}

\begin{figure}[tb]
    \centering
    \includegraphics[width=0.85\linewidth]{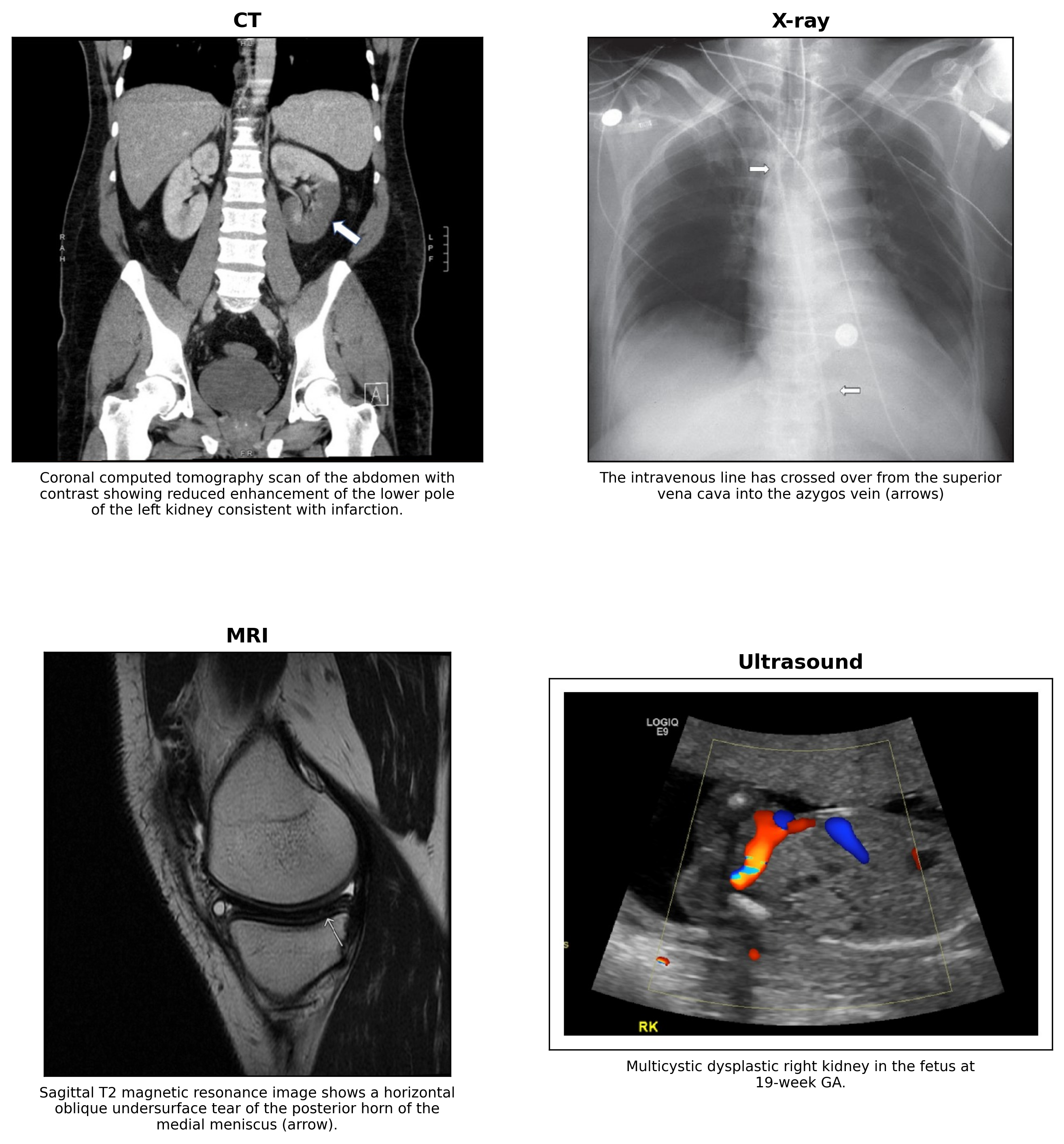}
    
    \caption{Representative ROCOv2 training examples drawn one per modality (CT, X-ray, MRI, Ultrasound) with their ground-truth captions. The dataset combines short modality-and-finding captions with longer clinically detailed descriptions, spans diverse anatomies even within a single modality, and is consistent in identifying the modality, body region, and salient findings.
    {Image credits} (all CC~BY): \emph{CT}~--~Muacevic et al.\ (\texttt{train\_55818}); \emph{X-ray}~--~Khan et al.\ (\texttt{train\_7504}); \emph{MRI}~--~Desai et al.\ (\texttt{train\_19872}); \emph{Ultrasound}~--~Mikuscheva et al.\ (\texttt{train\_58073}).}
    \label{fig:dataset images}
\end{figure}

The ImageCLEFmedical Caption 2026 task uses an updated and extended version of the ROCOv2 dataset ~\cite{ruckert2024rocov2}, a multimodal collection of radiology images (CT, X-ray, MRI, ultrasound, etc.) and their associated captions, extracted from the PMC Open Access subset.
Each caption can describe the modality, body part, and any findings or diseases visible in the image.
The original ROCOv2 dataset ~\cite{ruckert2024rocov2} contains 79,789 images across train/validation/test (59,958 / 9,904 / 9,927) with 1,947 unique CUIs.
The version released by the organizers expands this to 97,364 training images, 19,253 validation images, and 15,286 test images.
A few sample images are shown in Figure~\ref{fig:dataset images}.

\subsubsection{Dataset Exploration}

Before any modelling, we ran a set of exploratory analyses on the metadata to understand what we were dealing with.
Three things stood out and ended up shaping later decisions:
\begin{enumerate}
    \item {The CUI distribution is very long-tailed}:
    The training set contains 2,646 unique CUIs across 375,538 total occurrences, but the top 32 CUIs alone cover 50\% of all occurrences, while the 1,000th most frequent CUI shows up only 36 times.
    \item {Captions are short and bursty}:
    Most captions are between 8 and 40 words, but a small tail goes much longer.
    This matters for the caption-generation side because aggressive minimum-token settings or repetition penalties end up hurting more than they help on the short captions, which is consistent with how the generation hyperparameters were set.
    \item {Each image carries a small but variable number of CUIs}:
    Most images have 1–4 concepts, but a very small fraction of them has 10+, which is what makes this a true multi-label problem rather than something we can treat as classification.
\end{enumerate}

\subsubsection{BiomedCLIP for Dataset Structure}

Beyond the surface statistics, we wanted to know how much structure is already encoded in a strong pretrained medical vision encoder.
We use the BiomedCLIP ~\cite{zhang2023biomedclip} image embeddings (512-dim) as a fixed feature space and fit simple probes against the metadata to measure how easily modality and anatomy can be recovered, and which embedding dimensions are responsible for the same.

Defining modality labels:
The dataset does not ship with an explicit modality column, so we derive labels from the CUI list of each image.
We pick six CUIs that correspond to the dominant radiology modalities in ROCOv2 — CT, X-ray, MRI, Ultrasound, Angiography, and PET — and label each image with the most specific modality among those that appear in its CUIs.
Any image whose caption does not surface one of these six CUIs is put into a residual seventh class called Other.
This is a deliberate simplification of the full UMLS modality space, chosen so that each class has enough support for reliable probing.

\begin{figure}[tb]
    \centering
    \includegraphics[width=0.85\linewidth]{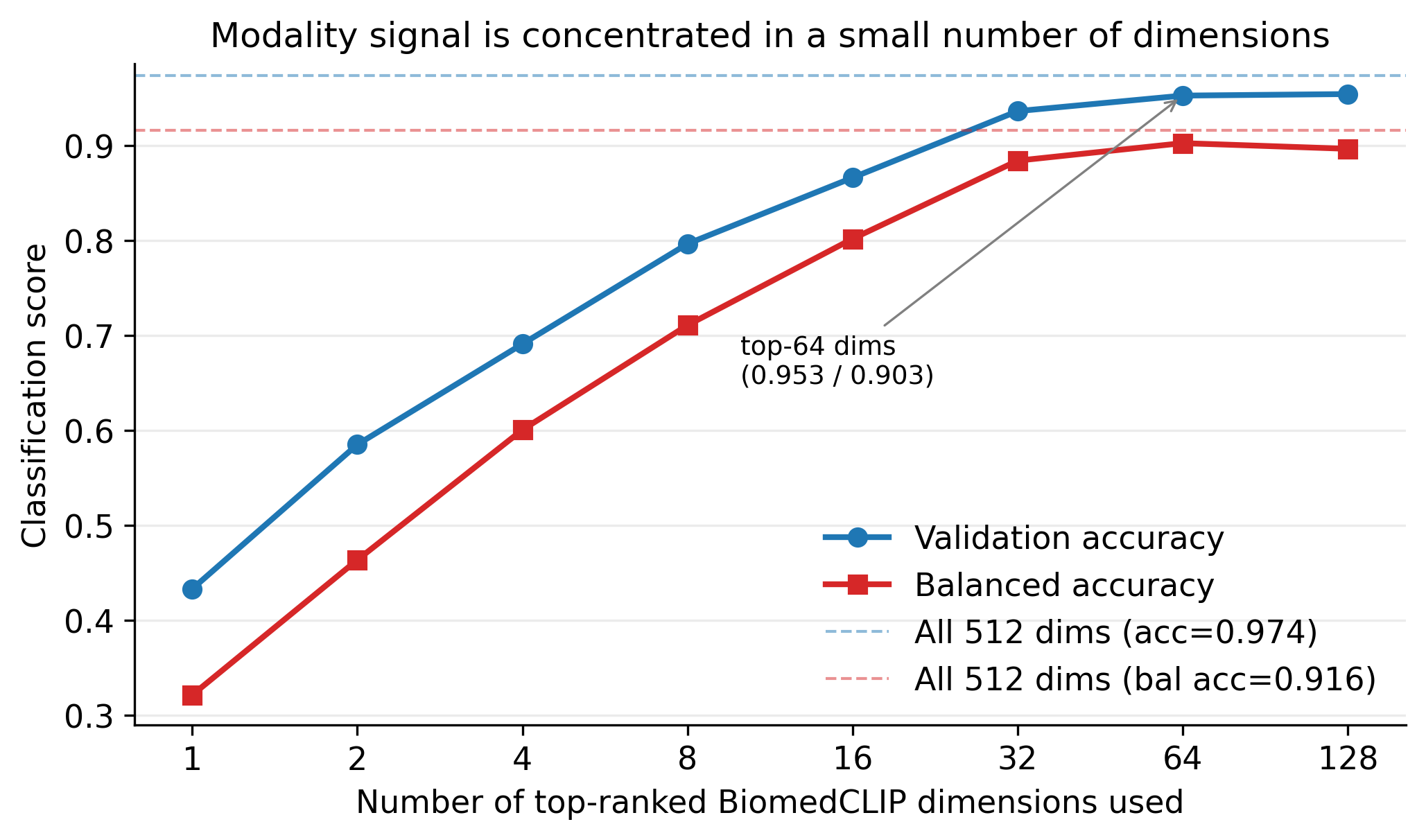}
    
    \caption{Modality classification score as a function of the number of top-ranked BiomedCLIP dimensions used as input to a fresh logistic regression. Dimensions are ranked once by feature importance from several all-512 classifiers; top-K then takes the K most-important dims under that ranking. 7-way = 6 dominant radiology modalities (CT, X-ray, MRI, Ultrasound, Angiography, PET) plus a residual Other class. Dashed lines show the full-512-dim reference.}
    \label{fig:modality signal}
\end{figure}

\begin{figure}[tb]
    \centering
    \includegraphics[width=0.85\linewidth]{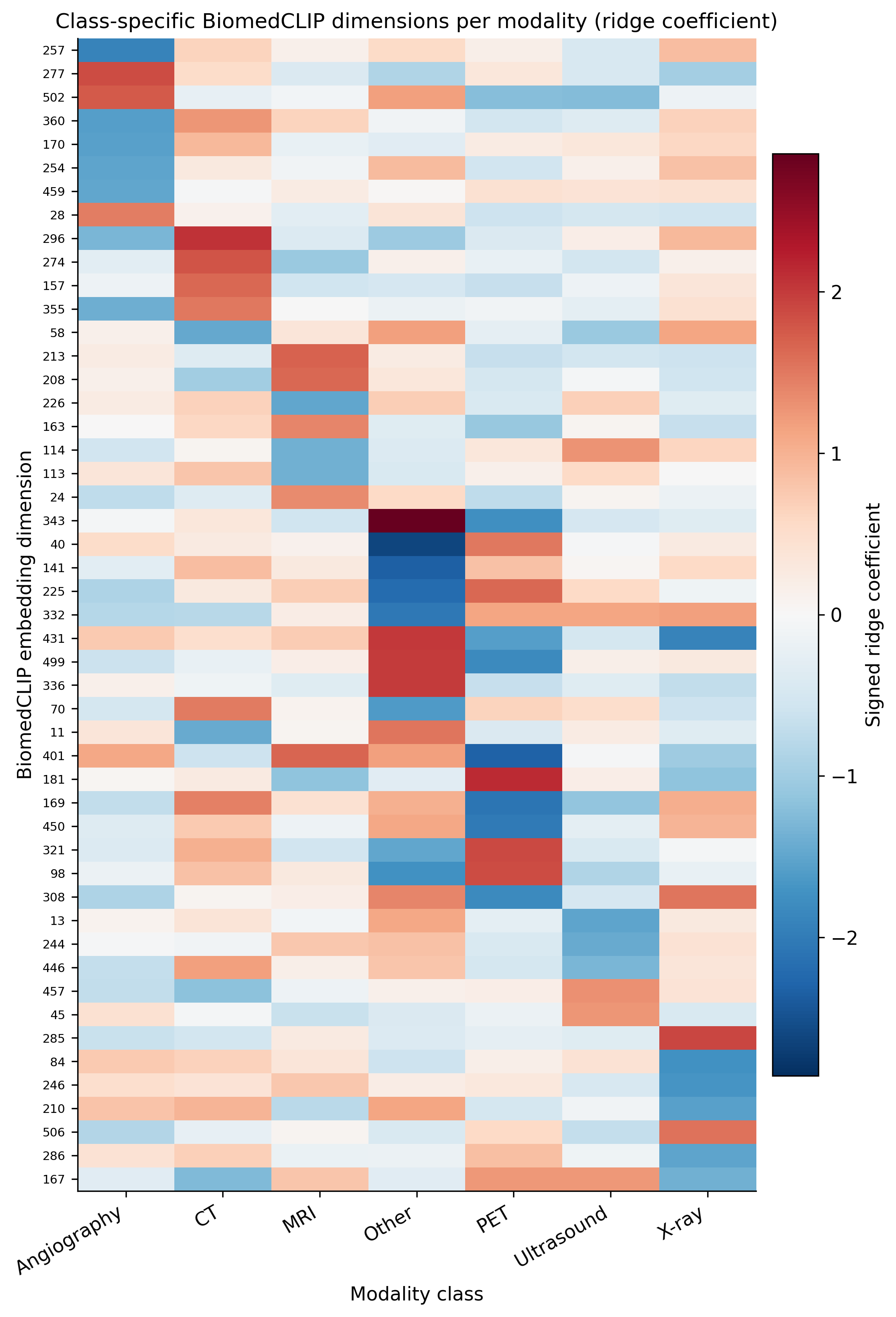}
    
    \caption{Per-class ridge coefficients of the top class-specific BiomedCLIP dimensions. Each cell is the signed weight assigned by a ridge linear classifier to dimension \textit{d} when predicting class \textit{c} — positive (red) means "this dim being high pushes predictions toward this class", negative (blue) means the opposite.}
    \label{fig:dim heatmap}
\end{figure}

\begin{figure}[tb]
    \centering
    \includegraphics[width=0.85\linewidth]{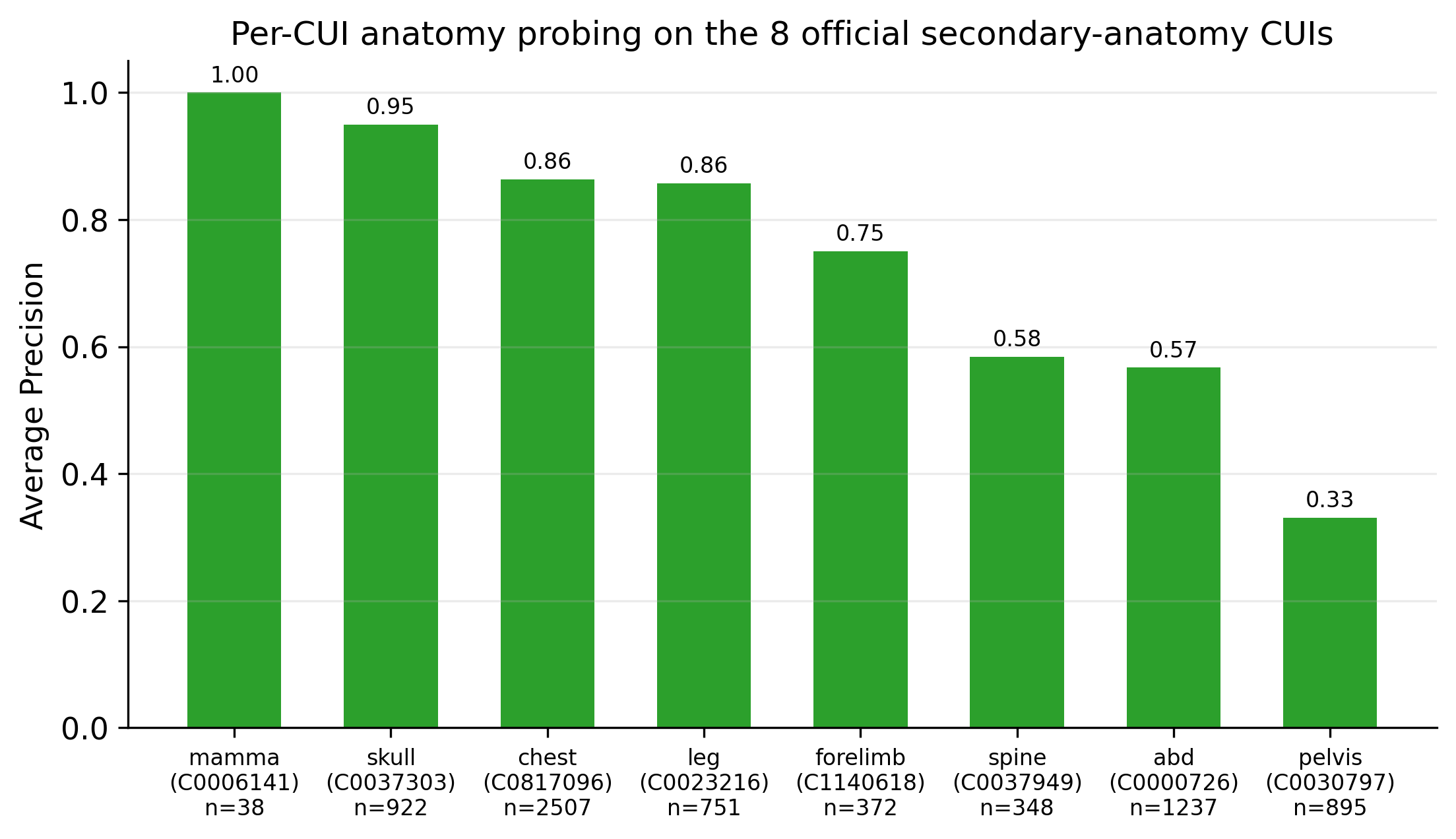}
    
    \caption{Per-CUI Average Precision (AP) for the eight official secondary-anatomy CUIs on the validation split, scored by a binary ridge probe trained on frozen BiomedCLIP image embeddings. For each CUI, the probe is asked one yes/no question — "does this image contain this anatomy?" — and outputs a real-valued score per image. AP summarizes how well that score ranks the positives above the negatives: we sort all validation images by score, walk down the list, and at each rank compute precision and recall against the ground truth. AP = 1.00 means every positive image is ranked above every negative one (perfect ranking); AP equal to the class prevalence means the ranking is no better than random. The number n under each bar is the count of validation images whose ground-truth CUI list contains that anatomy. Higher n with high AP = stronger evidence; mamma's perfect AP at n = 38 is encouraging but supported by far fewer examples than chest's AP = 0.86 at n = 2507.}
    \label{fig:anatomy analysis}
\end{figure}

\begin{figure}[tb]
    \centering
    \includegraphics[width=0.85\linewidth]{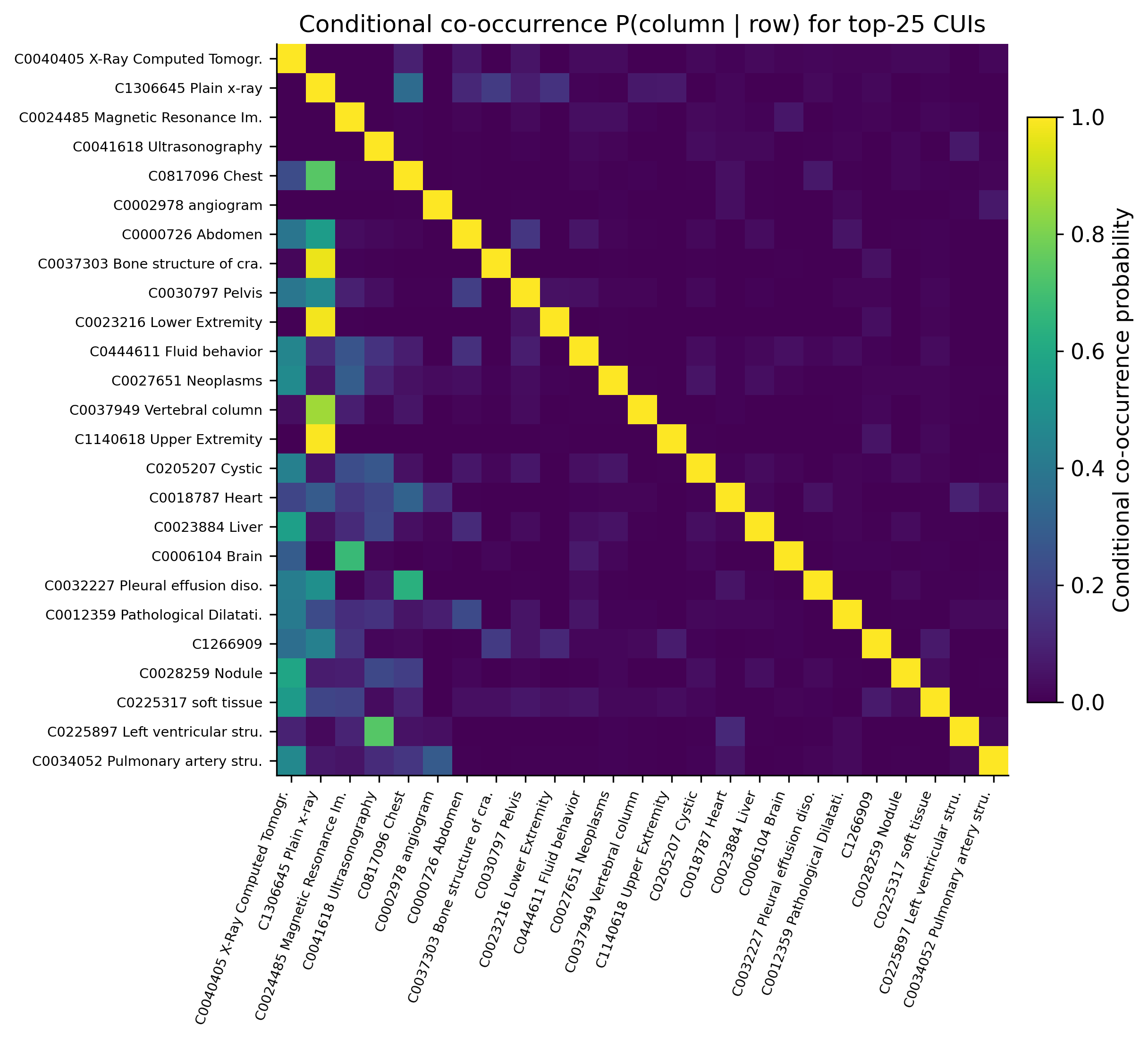}
    
    \caption{Conditional co-occurrence of the top-25 most frequent CUIs in the training set. Each cell shows how often the column's CUI appears in images that contain the row's CUI — so reading row-then-column tells you what concepts tend to accompany the row's CUI, while reading column-then-row reverses the conditioning, which makes the matrix non-symmetric. Bright off-diagonal patches reveal semantic clusters: modality CUIs anchor their own neighborhoods (X-ray $\to$ chest, CT $\to$ abdomen/pelvis, MRI $\to$ brain), and certain findings travel reliably with specific anatomies.}
    \label{fig:cui co-occurrence}
\end{figure}

\begin{enumerate}
    \item \textbf{Modality probing}:
    Using the modality CUIs in metadata.parquet as labels, we trained a series of light classifiers (L2/L1 logistic regression, XGBoost, cosine kNN, and a 2-layer MLP) on top of the frozen BiomedCLIP embeddings.
    The two takeaways from this sweep:
    \begin{itemize}
        \item BiomedCLIP already separates modalities extremely well.
        A simple 2-layer MLP reaches 0.9554 validation accuracy and an XGBoost classifier pushes accuracy to 0.9724.
        This is without any fine-tuning of the vision encoder.
        \item The signal is concentrated in a small number of dimensions.
        As Figure~\ref{fig:modality signal} shows, accuracy climbs very quickly with the number of top-ranked dimensions: top-1 dim only gives 0.43, top-8 already gives 0.80, top-32 gives 0.94, and top-64 gives 0.95 — essentially matching the full 512-dim model.
        Different modalities also "live" on different dimensions (Figure~\ref{fig:dim heatmap}).
    \end{itemize}
    \item \textbf{Anatomy probing}:
    Unlike modality, anatomy is multi-label — a single image can carry several anatomy CUIs — so we train one binary ridge probe per CUI rather than a single multiclass classifier.
    On the 8 official secondary-anatomy CUIs (chest, abdomen, skull, pelvis, leg, spine, forelimb, mamma), the all-512 probes give a per-image secondary-anatomy F1 of 0.7993, with compact top-64-dim probes still reaching 0.7827.
    Figure~\ref{fig:anatomy analysis} breaks this down per CUI: mamma and skull are essentially solved (AP = 1.00 and 0.95 respectively), chest and leg are strong, and pelvis is the hardest (AP = 0.33), which matches the intuition that pelvis images are visually heterogeneous and often overlap with abdomen and spine.
    \item \textbf{CUI co-occurrence}:
    Finally, we look at how the most frequent CUIs co-occur within the same image (Figure~\ref{fig:cui co-occurrence}).
    The co-occurrence heatmap shows clear semantic clusters — modality CUIs tend to occur within each cluster (e.g. CT routinely co-appears with abdomen/pelvis/chest concepts; X-ray with chest and skeletal concepts), and certain findings travel together in very predictable patterns.
    This is one of the reasons for attempting concept-anchored model: the label set is far from independent, and conditioning the caption model on a small set of anchor concepts could improve the results.

\end{enumerate}

\subsection{Fine-tuning Vision Language Model (Gemma-3)}

One approach we experimented with is fine-tuning an open-sourced Vision Language Model (VLM) to complete both Concept Detection and Caption Prediction tasks. The general structure of the fine-tuned VLM architecture is illustrated by Figure 6. A VLM takes an image as input and generates text as output, and has three core components: a vision encoder that extracts image features into embeddings, a language model that generates the text output, and a projection layer that maps the visual tokens into the same embedding space as the text tokens.

\begin{figure}[tb]
    \centering
    \includegraphics[width=0.85\linewidth]{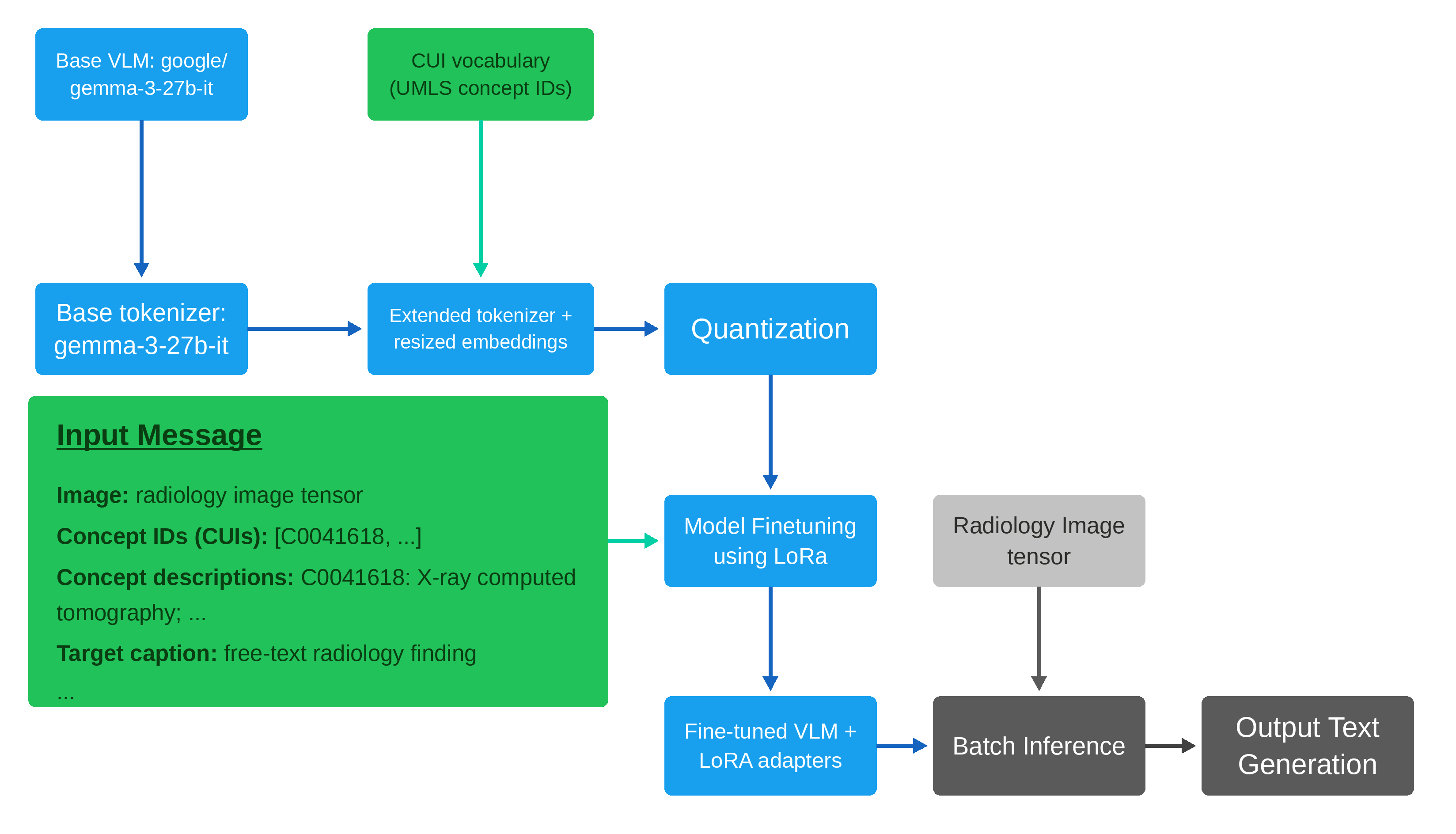}
    
    \caption{End-to-end multimodal fine-tuning architecture for Gemma-3-27B. The pipeline illustrates the expansion of the tokenizer vocabulary with UMLS Concept IDs and the application of 4-bit QLoRA adapters, allowing the model to jointly process image tensors and discrete medical concepts for unified caption generation.}
    \label{fig:vlm_architecture}
\end{figure}

\textbf{Fine-tuning with Quantized Low-rank Adaptation (QLoRA)}: QLoRA is a quantized variant of LoRA, an efficient fine-tuning method that freezes the base model weights and adds trainable adapters that learn task-specific updates. We implemented 4-bit quantization using the \texttt{BitsAndBytesConfig} class from the HuggingFace Transformers library. A common setup freezes the vision encoder and projection layer and adds LoRA adapters only to the attention modules of the language model. In contrast, we used the HuggingFace \texttt{peft} library to attach adapters to every \texttt{nn.Linear} layer in the model, enabling end-to-end multimodal fine-tuning of the vision encoder, connector, and language model simultaneously. The model was trained for 2 epochs with a batch size of 16. To fit the 27B-parameter model within GPU memory limits, we used Flash Attention-2, which maintained an effective batch size of 16 without exceeding VRAM. The learning rate was set to 7e-5 with a cosine-decay schedule and a 3\% warmup; we used a slightly lower rate than the 1e-4--2e-4 typical of smaller 7B/8B models to maintain training stability at this scale, and the warmup ramps the randomly initialized adapters in gradually to avoid an initial gradient spike.

\textbf{Model and Token Configuration:} We started from a publicly available Gemma 27B VLM checkpoint from Hugging Face. The original tokenizer does not contain medical CUIs, so we extended it by adding all the CUI codes (e.g., “C0041618”) as special tokens. The embedding matrix and output layer were resized accordingly so that the model could assign a distinct representation and prediction logit to every CUI. For each training example, we constructed a text prompt that interweaves the image with both human-readable descriptions and the discrete CUI codes. Intuitively, the description gives rich semantic context while the CUI token provides a compact, standardized label. By always showing CUIs together with their descriptions during fine-tuning, we encourage the model to align the discrete codes (CUIs) and captions with the underlying visual patterns. Figure 7 shows some examples of the concepts and captions generated by the fine-tuned VLM compared with the ground truths. The fine-tuned Gemma-3 model successfully bridges localized visual features (such as the indicated lesion) with their precise UMLS concept codes while maintaining accuracy and fluency in the generated clinical caption.

\begin{figure}[tb]
    \centering
    \includegraphics[width=0.85\linewidth]{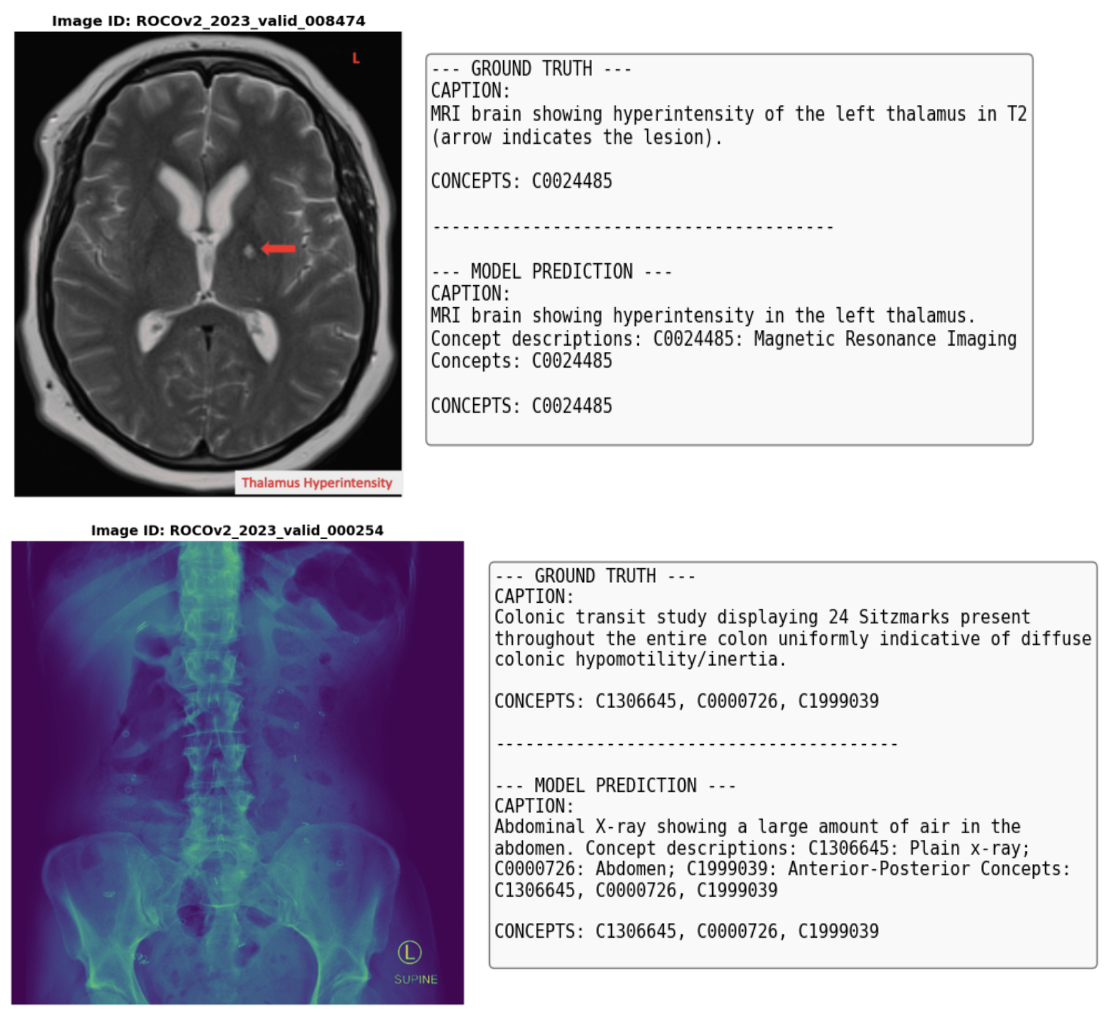}
    
    \caption{Qualitative comparison of ground truth and model-predicted outputs from the fine-tuned Gemma-3-27B model. The examples demonstrate the model's ability to accurately identify target concepts (e.g., C0024485: Magnetic Resonance Imaging) and generate coherent free-text findings that align closely with the ground-truth annotations.Image credits (all CC~BY): Rückert et al.\ (\texttt{valid\_008474}, \texttt{valid\_000254})}
    \label{fig:vlm_examples}
\end{figure}

\subsection{Concept Detection}

\subsubsection{VLM Inference for Concept Extraction}

The base model (google/gemma-3-27b-it) is loaded in 4-bit NormalFloat (nf4) precision. To prevent accuracy degradation during generation, the computation data type is set to bfloat16. The trained LoRA adapter is injected into the frozen 4-bit base model using the peft library. The AutoProcessor is used to handle both the tokenization of the text prompt and the pixel-value transformation of the medical scans. In terms of prompting, the system prompt designates the model as “a digital radiologist who can understand the medical scan of images”, and the user prompt instructs the model to “return the concepts and their descriptions, only the concepts are extracted”. Both the system prompt and the user prompt are passed alongside the image tensor. Crucially, the inference strategy for the concept detection task disables sampling (which traditionally relies on temperature and Top-K/Top-p to introduce variance). By mathematically forcing the model into a purely deterministic path—equivalent to greedy decoding—the system always selects the highest-probability tokens. This strict decoding strategy actively suppresses hallucinations, preventing the model from outputting invalid CUI codes.

\subsubsection{Three-way Vision Ensemble and Honest Threshold Tuning}
\label{sec:three-way}

\begin{figure}[tb]
    \centering
    \includegraphics[width=0.85\linewidth]{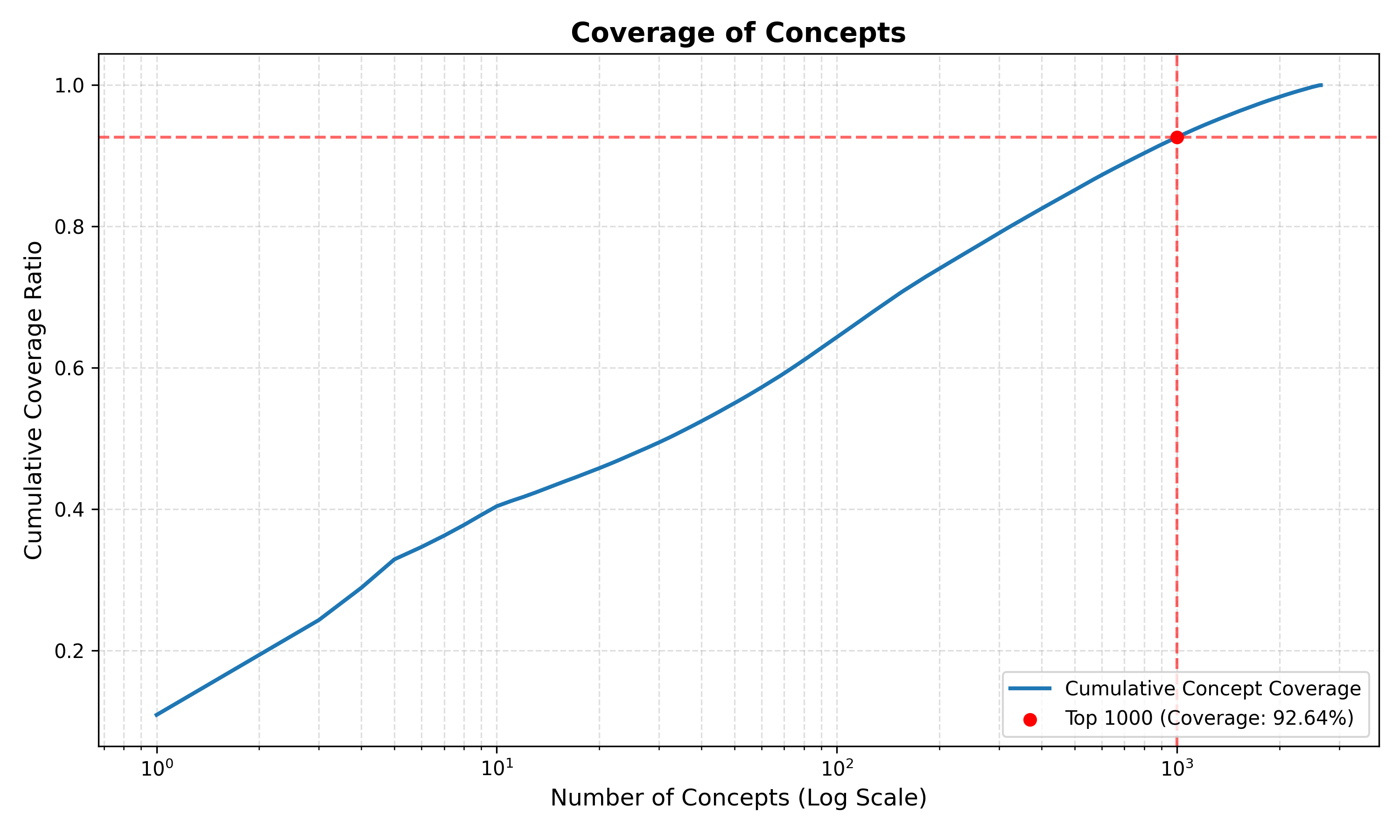}
    
    \caption{Cumulative coverage of concept occurrences as a function of the number of most-frequent CUIs included.}
    \label{fig:concept_coverage}
\end{figure}

\textbf{Label Space Selection:} The full concept vocabulary in our training data comprises 2{,}646 unique CUIs spanning 375{,}538 total occurrences. The labels exhibit a severe long-tail pattern: the top 32 CUIs account for 50\% of all occurrences, while the 1{,}000th most frequent CUI appears only 36 times. Because reliable training of a multi-label classifier requires sufficient positive examples per class, we restricted the label space to the top 1{,}000 most frequent CUIs. As Figure~\ref{fig:concept_coverage} shows, this subset covers 92.64\% of all concept occurrences in the training data, providing a highly stable training signal while intentionally sacrificing recall on rare concepts to protect overall classifier calibration.

\textbf{Baseline and Test-Time Augmentation:} To establish a baseline, we trained a dual-backbone CNN ensemble (DenseNet-121~\cite{huang2017densenet} + EfficientNet-B0~\cite{tan2019efficientnet}). Evaluated on the validation set, the unaugmented baseline achieved an $F_1$ score of 0.4789. Applying data augmentation during training and Test-Time Augmentation (TTA) during inference raised the score to 0.5095. For TTA, we generated augmented views using a small positive rotation (+5$^\circ$), a small negative rotation ($-$5$^\circ$), and a slight center zoom (1.05$\times$). This TTA strategy was adopted for all subsequent models.

\textbf{Stand-Alone Encoders and Ensemble Weights:} We trained three architecturally distinct vision encoders on the top-1000 CUI label space:
\begin{enumerate}
    \item {ConvNeXt-V2-Base} (89.0M params): Fine-tuned with Asymmetric Loss (ASL)~\cite{benbaruch2021asymmetricloss} to down-weight well-classified negative samples. Validation $F_1$ = 0.5923.
    \item {BiomedCLIP ViT-B/16} (86.6M params): Leveraged strong medical visual priors from PubMed Central pretraining~\cite{zhang2023biomedclip}. Validation $F_1$ = 0.5889.
    \item {DenseNet-169} (14.1M params): Provided vital architectural diversity at a fraction of the parameter count~\cite{huang2017densenet}. Validation $F_1$ = 0.5889.
\end{enumerate}

To maximize performance, we created a late-fusion weighted-average ensemble, assigning fixed weights of 0.60 to ConvNeXt-V2, 0.20 to BiomedCLIP, and 0.20 to DenseNet-169. These weights were determined by coordinate ascent on the validation set. The three-way ensemble achieved a validation $F_1$ of 0.6061 prior to threshold tuning. An experimental dual-stream cross-attention model combining ConvNeXt and BiomedCLIP was also tested (validation $F_1$ = 0.5906) but did not surpass this late-fusion approach, suggesting that for this task, decoupled backbones with late fusion are at least as effective as joint multi-modal fine-tuning.

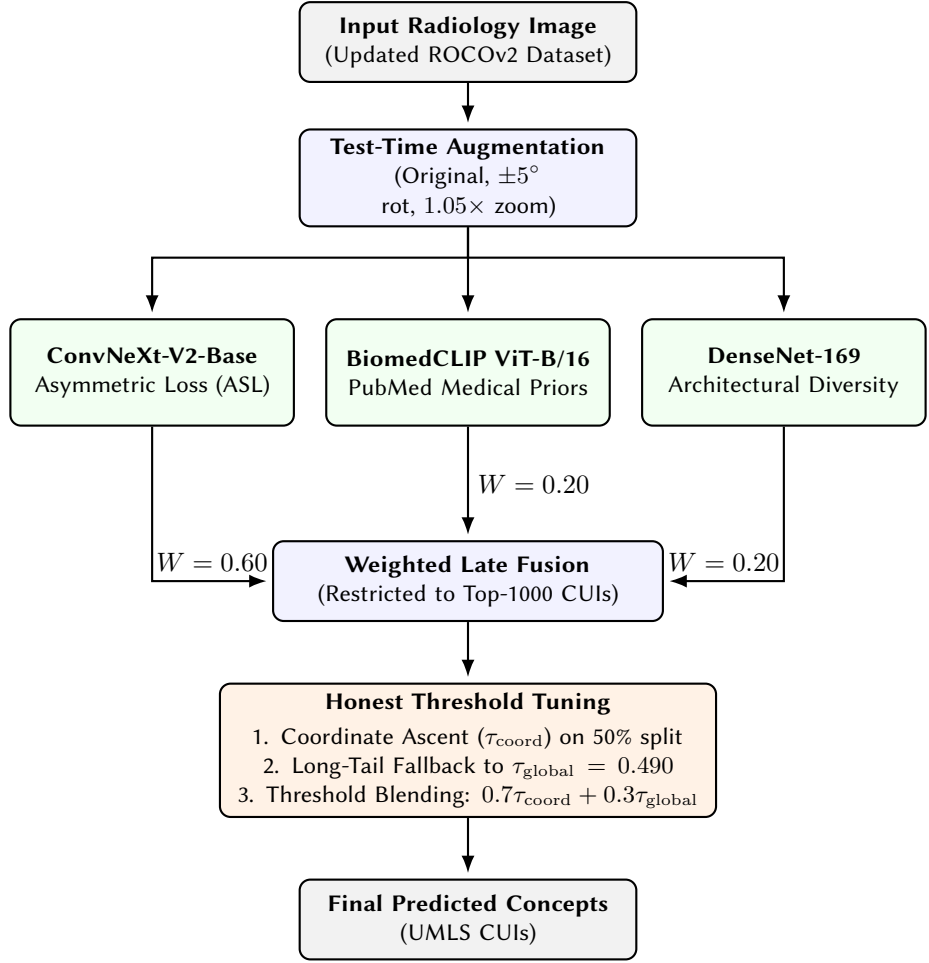
\begin{figure}[tb]
\centering
\begin{tikzpicture}[
    auto,
    block/.style = {rectangle, draw, thick, fill=blue!5, text width=12em, text centered, rounded corners, minimum height=3em, font=\footnotesize},
    model/.style = {rectangle, draw, thick, fill=green!5, text width=10em, text centered, rounded corners, minimum height=4em, font=\footnotesize},
    highlight/.style = {rectangle, draw, thick, fill=orange!10, text width=18em, text centered, rounded corners, minimum height=4em, font=\footnotesize},
    line/.style = {draw, thick, -Latex, shorten >=2pt},
]

\node [block, fill=gray!10] (input) {\textbf{Input Radiology Image}\\(Updated ROCOv2 Dataset)};

\node [block, below=0.6cm of input] (tta) {\textbf{Test-Time Augmentation}\\(Original, $\pm 5^\circ$ rot, $1.05\times$ zoom)};

\node [model, below=1.2cm of tta] (biomed) {\textbf{BiomedCLIP ViT-B/16}\\ PubMed Medical Priors};
\node [model, left=0.4cm of biomed] (convnext) {\textbf{ConvNeXt-V2-Base}\\ Asymmetric Loss (ASL)};
\node [model, right=0.4cm of biomed] (densenet) {\textbf{DenseNet-169}\\ Architectural Diversity};

\node [block, below=1.5cm of biomed, text width=14em] (fusion) {\textbf{Weighted Late Fusion}\\ (Restricted to Top-1000 CUIs)};

\node [highlight, below=0.8cm of fusion] (threshold) {\textbf{Honest Threshold Tuning}\\ \vspace{2pt} 1. Coordinate Ascent ($\tau_{\mathrm{coord}}$) on 50\% split\\ 2. Long-Tail Fallback to $\tau_{\mathrm{global}} = 0.490$\\ 3. Threshold Blending: $0.7\tau_{\mathrm{coord}} + 0.3\tau_{\mathrm{global}}$};

\node [block, fill=gray!10, below=0.8cm of threshold] (output) {\textbf{Final Predicted Concepts}\\(UMLS CUIs)};

\path [line] (input) -- (tta);

\path [line] (tta.south) -- ++(0,-0.4) -| (convnext.north);
\path [line] (tta.south) -- (biomed.north);
\path [line] (tta.south) -- ++(0,-0.4) -| (densenet.north);

\path [line] (convnext.south) |- node[above, pos=0.75] {$W=0.60$} (fusion.west);
\path [line] (biomed.south) -- node[right] {$W=0.20$} (fusion.north);
\path [line] (densenet.south) |- node[above, pos=0.75] {$W=0.20$} (fusion.east);

\path [line] (fusion) -- (threshold);
\path [line] (threshold) -- (output);

\end{tikzpicture}
\caption{Architecture of our top-performing three-way ensemble pipeline for Concept Detection (Task 1). The system utilizes Test-Time Augmentation, weighted late fusion of architecturally diverse vision encoders, and regularized Honest Threshold Tuning to prevent long-tail overfitting.}
\label{fig:task1_architecture}
\end{figure}

\textbf{Honest Threshold Tuning and Regularization:} Rather than applying a single global decision threshold, we optimized a per-concept threshold $\tau_c$ for each CUI. Traditional coordinate ascent on the full validation set led to severe overfitting on rare concepts, yielding an artificially inflated expected score. To construct a submission that would generalize to the blind test set, we engineered an ``Honest Threshold Tuning'' pipeline:
\begin{enumerate}
    \item {Deterministic Split:} The 19{,}253 validation images were split 50/50. Coordinate ascent was performed strictly on the tuning half (9{,}620 images), following the approach of Chatzipapadopoulou et al.~\cite{auebnlpcaption2025}.
    \item {Long-Tail Fallback:} For the 910 concepts with fewer than 50 occurrences in the tuning split, the coordinate-ascent thresholds were discarded and reverted to a safe global baseline ($\tau_{\mathrm{global}} = 0.490$).
    \item {Threshold Blending:} To pull extreme thresholds back toward a safe center, we blended the remaining coordinate-ascent thresholds: $\tau_{\mathrm{final}} = 0.7\,\tau_{\mathrm{coord}} + 0.3\,\tau_{\mathrm{global}}$.
\end{enumerate}
Evaluated on the held-out verification half, this regularized approach provided a highly realistic performance estimate, successfully closing the generalization gap prior to official test set submission.
Figure~\ref{fig:task1_architecture} shows our architecture for Task~1.

\subsubsection{BiomedCLIP Retrieval Pipeline}
\label{sec:bmc-retrieval}
Complementary to the CNN ensemble in Section 3.3.2 and the VLM extractor in Section 3.3.1, we explored a much lighter, training-free pipeline built entirely on top of frozen BiomedCLIP image embeddings.
The motivating intuition follows directly from our dataset analysis in Section 3.1.2: if BiomedCLIP's pretraining already groups visually and semantically similar radiology images, then for a new image its nearest training neighbors should carry most of the relevant CUIs already.
The best configuration from this line was used as one of our official Task 1 submissions.

\textbf{KNN Retrieval over BiomedCLIP Embeddings.}
We $L_2$-normalize the 512-dim BiomedCLIP embeddings for the train, validation, and test splits, and for each query image retrieve the top-$K$ nearest training images by cosine similarity.
The CUIs carried by those neighbors were then aggregated.
After sweeping $K \in \{1, 3, 5, 10, 20, 50\}$ and seven aggregation strategies, the strongest configuration was \emph{$K = 50$ with weighted-ratio aggregation at ratio $0.35$}: for each candidate CUI we sum the cosine similarities of the neighbors that carry it and keep the CUI only if that sum is at least $35\%$ of the maximum possible.
This decoder is deliberately precision-oriented and predicts roughly $1.4$ CUIs per image on average---well below the dataset mean---which is well suited to the official per-image $F_1$ metric's penalty on large predicted sets.

\textbf{Manual-CUI Retuning.}
Because the official secondary score is computed against only 15 manually curated CUIs, the primary and secondary metrics can be tuned somewhat independently.
We added a second KNN pass restricted to those 15 CUIs and merged its predictions into the base configuration.
The best variant was a top-1 decode at $K = 20$, unioned with the base prediction.

We submitted this configuration---KNN $K = 50$ with weighted-ratio $0.35$, unioned with a $K = 20$ top-1 manual-CUI retune-- as one of our official Task~1 runs.
The test-set scores are reported in Results section.

\subsection{Caption Prediction}

\subsubsection{VLM Inferencing for Caption Prediction}
\label{sec:gemma-3}
For Caption Prediction, we also utilized the same fine-tuned Gemma-3 27B model and QLoRA adapter established in Section 3.2. We implemented the same underlying Gemma-3-27B model, quantization setup, and LoRA adapter. However, there are differences in how inference is conducted. For each batch, images are opened via PIL and converted to RGB. Max length of token is set to 150 to ensure the model has enough room to fully describe the image while strictly preventing it from "hallucinating" endlessly. No minimum length of tokens is set because medical captions can sometimes be extremely brief (e.g., "Normal chest X-ray."). Forcing a minimum token length would cause the model to artificially pad its answers. There is no repetition penalty because medical terminology must be precise and is frequently repeated (e.g., "lesion," "artery"). Applying a repetition penalty can accidentally force the model to use inaccurate synonyms. Additionally, unlike concept inferencing, caption inferencing uses beam search (3 beams), which ensures that the outputs are mathematically stable and significantly reduces the risk of infinitely repeating loops. 

\subsubsection{Concept Anchoring and VLM Ablation (InstructBLIP, LLaVA, BLIP)}
\label{sec:concept_anchoring}

\textbf{Concept Anchoring:} A core idea explored in our submissions is \emph{concept anchoring}: injecting predicted UMLS concepts from Task~1 into the prompt at training and inference time as structured text, organized into three semantic buckets (Modality, Anatomy, Findings) determined by the UMLS Type Unique Identifier (TUI) of each concept. For example, an image associated with CUIs for CT scan, chest, and pneumonia is presented to the model as:
\begin{quote}
\texttt{Modality: CT scan. Anatomy: chest imaging. Findings: pneumonia.}
\end{quote}
During training, we use the ground-truth CUIs from \texttt{concepts.csv}. At inference, we apply a ``dense-40'' strategy: a flat 0.40 probability cutoff on our Task~1 ensemble predictions, yielding an average of 2.05 concepts per image as factual anchors.

\textbf{Vision-Language Architectures:} We compared three architectures, fine-tuning each with and without concept anchoring:
\begin{enumerate}
    \item {InstructBLIP-Flan-T5-XL}~\cite{dai2023instructblip}: Combines a frozen ViT-G/14 encoder, a Q-Former bridge, and Flan-T5-XL~\cite{chung2024scaling}. Fine-tuned via LoRA~\cite{hu2022lora} (rank~16) on the language model ($\sim$9.4M trained parameters) for 3 epochs in FP32.
    \item {LLaVA-Med v1.5 Mistral-7B}~\cite{li2023llavamed}: Combines a CLIP ViT-L/14 encoder with Mistral-7B, pretrained on PubMed Central biomedical figure-caption pairs. Fine-tuned via LoRA ($\sim$28M trained parameters) for 3 epochs.
    \item {BLIP-base}~\cite{li2022blip}: Couples a ViT-Base encoder with a BERT-base text decoder. Fully fine-tuned for 8 epochs, following the recipe of UMUTeam~\cite{umuteam2025} from the 2025 task. During evaluation, we observed that decoder-only models truncated outputs when processing long concept anchors. By expanding the generation limit from 50 to 80 tokens, we recovered substantial gains in fluency metrics.
\end{enumerate}

\textbf{Caption Merging Strategy:} For BLIP, where concept anchoring's relevance penalty roughly cancelled its factuality gain, we
applied a custom ``Vizwins'' merging algorithm. For each image, the rule engine selected the anchored caption if it contained strictly more concept anchor hits than the visual-only caption; if tied on concept hits, it defaulted to the visual-only caption to preserve grammatical fluency.

\subsubsection{MedGemma-4B Zero-Shot with PMC-Legend Prompting}

In parallel with the Gemma-3 27B pipeline, we ran experiments on the 4B-parameter MedGemma model—pretrained on PubMed Central data and therefore well-matched to ROCOv2's caption style.
Our strongest configuration was zero-shot prompted inference; fine-tuning ablations and a post-hoc reranker did not surpass it on validation.
Quantitative results for all variants are reported in Section~\ref{sec:caption-results}.

\textbf{MedGemma-4B Zero-Shot with the PMC-Legend Prompt.}
\label{sec:medgemma-zsp}
We prompt the base MedGemma-4B-it checkpoint with a system message framed in the style of a PubMed Central figure legend, asking the model to name the imaging modality, anatomical region, and notable findings in one to two sentences.
Decoding is deterministic greedy, and every output passes through a regex-based post-processor that strips markdown artifacts and collapses multi-line responses.
We swept five prompt variants on the validation set; the PMC-legend framing consistently won on ROUGE-1 and BLEURT, and was used for the official submission.

\textbf{Fine-Tuning Ablation: QLoRA, Full SFT, and Joint Concept\,+\,Caption SFT.}
We ran three fine-tuning variants on the ROCOv2 training captions.
The QLoRA baseline applies rank-16 LoRA adapters on the attention projections; a second variant also targets the MLP projections.
Full SFT updates all parameters at a low learning rate to balance adaptation with preservation of pretrained knowledge.
The Joint SFT variant changes the training target to a two-part block --- \texttt{CONCEPTS: <UMLS term names>\textbackslash nCAPTION: <caption text>} ---  so that the model conditions caption generation on the concepts it just predicted, and also yields a free Task 1 prediction noted at the end of Section~\ref{sec:bmc-retrieval}.
All three fine-tunes underperformed zero-shot, with BLEURT dropping sharply; per-variant numbers are in Section~\ref{sec:caption-results}.

\textbf{BiomedCLIP Cross-Modal Retrieval and Reference-Free Reranker (Post-hoc).}
After the official submission, we explored combining multiple candidate captions per image and selecting the best without references.
Candidates came from three sources: BiomedCLIP nearest-neighbor retrieval, an optional ridge-regression-aligned retrieval variant, and few-shot RAG captions from MedGemma.
A reference-free selector scores each candidate by BiomedCLIP image–caption similarity plus a per-source bias and a soft length penalty.
This post-hoc configuration was not submitted; results are reported in Section~\ref{sec:caption-results}.

\section{Results}

\subsection{Concept Detection Results}
\label{sec:concept-results}

\begin{table}[t]
\centering
\caption{Concept Detection results on the validation and official test sets.}
\label{tab:task1_dev_results}
\begin{tabular}{lcc}
\toprule
\textbf{Model} & \textbf{Primary F1} & \textbf{Notes} \\
\midrule
Baseline (DenseNet-121 + EfficientNet-B0) & 0.4789 & no aug/TTA \\
Fine-tuned Gemma-3                      & 0.4893 & 27B parameters, 2 epochs  \\
Baseline + TTA                      & 0.5095 & test-time augmentation  \\
ConvNeXt-V2-Base                          & 0.5923 & 89.0M params, ASL loss \\
BiomedCLIP ViT-B/16                       & 0.5889 & 86.6M params \\
DenseNet-169                              & 0.5889 & 14.1M params \\
Dual-Stream (ConvNeXt + BiomedCLIP)       & 0.5906 & cross-attention fusion \\
2-Way Ensemble (ConvNeXt + BiomedCLIP)    & 0.5941 & weighted average \\
Three-way Ensemble                   & 0.6061 & ConvNeXt + BiomedCLIP + DenseNet \\
BiomedCLIP KNN Retrieval   & 0.5914 & K=50, weighted-ratio 0.35 \\
BiomedCLIP KNN + Manual Retune   & 0.5920 & K=20 top-1 manual retune \\
Three-way Ensemble (Test set)                 & 0.5763 & raw coordinate ascent \\
Three-way Ensemble (Test set)                 & 0.5768 & majority-vote \\
\textbf{Three-way Ensemble (Test set)}                   & \textbf{0.5790} & Honest Threshold Tuning \\
BiomedCLIP KNN + Manual Retune (Test set)   & 0.5780 & retrieval-only, K=50 + manual retune \\
\bottomrule
\end{tabular}
\end{table}

\begin{table}[t]
\centering
\caption{Top participant submissions on the hidden test set for the 2026 Concept Detection task.}
\label{tab:concept_detection_leaderboard}
\begin{tabular}{llcc}
\toprule
\textbf{Rank} & \textbf{Participant} & \textbf{F1 (primary)} & \textbf{F1 (secondary)} \\
\midrule
\textbf{1} & \textbf{dsgt\_caption} & \textbf{0.5790} & \textbf{0.9657} \\
2 & archimedes & 0.5781 & 0.9591 \\
3 & quocthai91206 & 0.5725 & 0.9533 \\
4 & basharderar & 0.5725 & 0.9419 \\
5 & georkarandreas & 0.5721 & 0.9503 \\
6 & nhanbui & 0.5713 & 0.9553 \\
7 & bahareh0281 & 0.5648 & 0.9336 \\
8 & dingglebellw & 0.5606 & 0.9428 \\
9 & mihail\_tatomir & 0.5594 & 0.9465 \\
10 & krishnatewari & 0.5234 & 0.8403 \\
11 & tanduy2702 & 0.2739 & 0.9349 \\
\bottomrule
\end{tabular}
\end{table}

Table~\ref{tab:task1_dev_results} summarizes the progression of our model performance on the validation set and our official test set submissions. Starting from a dual-backbone CNN baseline ($F_1$ = 0.4789), the introduction of training augmentations and Test-Time Augmentation (TTA) yielded a substantial +0.0306 improvement. Among the standalone vision encoders, ConvNeXt-V2-Base achieved the highest individual validation score (0.5923). While an experimental dual-stream cross-attention approach achieved 0.5906, late-fusion ensemble proved more effective. Combining ConvNeXt and BiomedCLIP into a 2-way ensemble raised the score to 0.5941, and incorporating the architectural diversity of DenseNet-169 into our final three-way ensemble pushed our validation performance to 0.6061.

For our official test set submissions, our three-way ensemble with Honest Threshold Tuning, describe in Section~\ref{sec:three-way}, achieved a primary $F_1$ score of 0.5790 and a secondary $F_1$ of 0.9657, ranking 1st among all participating teams in the Concept Detection task. To isolate the impact of our thresholding strategy, we also submitted the same three-way ensemble using raw coordinate ascent thresholds, which achieved a lower score of 0.5763. A secondary configuration utilizing a strict majority-vote consensus scored 0.5768. The measurable improvement from 0.5763 to our winning 0.5790 confirms that our regularized, blended threshold strategy successfully mitigates the long-tail overfitting inherent to traditional coordinate ascent, preserving recall on difficult examples significantly better than baseline approaches.

Our complementary BiomedCLIP-retrieval submission, described in Section~\ref{sec:bmc-retrieval}, also reached the test set.
On the validation split, the base KNN configuration ($K=50$, weighted-ratio $0.35$) scored primary $F_1 = 0.5914$ and secondary $F_1 = 0.9282$; adding the $K=20$ top-1 manual-CUI retune lifted these to $0.5920\,/\,0.9290$---a small but consistent gain on both axes.
We submitted this final combined configuration as one of our official Task~1 runs.

On the held-out test set, the retrieval submission scored a primary $F_1$ of $0.5780$ and a secondary $F_1$ of $0.9599$.
The primary score is essentially tied with the three-way ensemble's $0.5790$, a $0.001$ gap, despite the retrieval pipeline using no fine-tuning and only a single pretrained vision encoder.
On the secondary 15-CUI track the ensemble retains a clear advantage ($0.9657$ vs.\ $0.9599$); we attribute this to the ensemble's per-concept threshold tuning, which has more room to optimize precision on the narrow manually-curated CUI subset than a retrieval-based decoder.

\subsection{Caption Prediction Results}
\label{sec:caption-results}

\begin{table}[t]
\centering
\small
\caption{Caption Prediction results on the validation and official test sets. Rel.~= Relevance average;
Fact.~= Factuality average; Overall = mean of the two.}
\label{tab:task2_results}
\begin{tabular}{lcccc}
\toprule
\textbf{Model / Configuration} & \textbf{Overall} & \textbf{Rel.} & \textbf{Fact.} & \textbf{Notes} \\
\midrule
\multicolumn{5}{l}{\textit{Gemma-3 (full fine-tune, $\sim$27B parameters, 150 tokens)}} \\
\quad 2 Beams (1.5 epochs) (Test set)         & 0.3446 & 0.5247 & 0.1644 & best epoch \\
\quad \textbf{3 Beams (2 epochs) (Test set)}         & \textbf{0.3571} & \textbf{0.5365} & \textbf{0.1777} & best epoch \\
\midrule
\multicolumn{5}{l}{\textit{InstructBLIP-Flan-T5-XL (LoRA, $\sim$9.4M trained)}} \\
\quad Visual-only      & 0.2879 & 0.4607 & 0.1151 & 3 epochs \\
\quad Concept-anchored & 0.3156 & 0.4692 & 0.1620 & dense-40 \\
\midrule
\multicolumn{5}{l}{\textit{LLaVA-Med v1.5 Mistral-7B (LoRA, $\sim$28M trained)}} \\
\quad Visual-only           & 0.3202 & 0.5148 & 0.1257 & 3 epochs \\
\quad Concept-anchored      & 0.3302 & 0.5107 & 0.1498 & dense-40 \\
\quad Concept-anchored (Test set)  & 0.3324 & 0.5134 & 0.1515 & dense-40 \\
\midrule
\multicolumn{5}{l}{\textit{BLIP-base (full fine-tune, $\sim$250M trained, 80 tokens)}} \\
\quad Visual-only (epoch 7)          & 0.3499 & 0.5328 & 0.1671 & best epoch \\
\quad Concept-anchored (epoch 5)     & 0.3476 & 0.5069 & 0.1884 & best epoch \\
\quad Merged Pipeline (Test set) & 0.3564 & 0.5351 & 0.1776 & Vizwins merging \\
\midrule
\multicolumn{4}{l}{\emph{MedGemma-4B (zero-shot, PMC-legend prompt, $\sim$4B parameters, no fine-tuning)}} \\
Greedy decoding (Test set) & 0.3186 & 0.5120 & 0.1253 & PMC-leg prompt-only \\
\bottomrule
\end{tabular}
\end{table}

\begin{table}[t]
\centering
\caption{Top participant submissions on the hidden test set for the 2026 Caption Prediction task.}
\label{tab:caption_prediction_leaderboard}
\begin{tabular}{llccc}
\toprule
\textbf{Rank} & \textbf{Participant} & \textbf{Overall} & \textbf{Relevance} & \textbf{Factuality} \\
\midrule
1 & heeseung & 0.3755 & 0.5786 & 0.1724 \\
2 & mendesheitor & 0.3603 & 0.5527 & 0.1679 \\
\textbf{3} & \textbf{dsgt\_caption} & \textbf{0.3571} & \textbf{0.5365} & \textbf{0.1777} \\
4 & georkarandreas & 0.3516 & 0.5288 & 0.1743 \\
5 & csmorgan & 0.3458 & 0.5377 & 0.1539 \\
6 & gfreitas & 0.3343 & 0.5201 & 0.1484 \\
7 & tanduy2702 & 0.3193 & 0.4999 & 0.1388 \\
8 & mihail\_tatomir & 0.3065 & 0.4705 & 0.1426 \\
9 & ronghaopan & 0.2990 & 0.4798 & 0.1181 \\
10 & uit-01100010en & 0.2814 & 0.4526 & 0.1102 \\
11 & krishnatewari & 0.2685 & 0.4346 & 0.1025 \\
12 & sebastian & 0.2673 & 0.4275 & 0.1072 \\
13 & nguyenthinh & 0.2527 & 0.4373 & 0.0680 \\
\bottomrule
\end{tabular}
\end{table}

Table~\ref{tab:task2_results} presents the Caption Prediction results on the validation and official test sets. For our fine-tuned Vision-Language Model pipeline, described in Section~\ref{sec:gemma-3}, the Gemma-3 27B model emerged as our strongest overall system. We evaluated different training durations and decoding strategies, finding that a two-epoch fine-tune paired with a 3-beam search yielded the best performance, outperforming the 1.5-epoch, 2-beam variant. On the held-out test set, this optimal configuration achieved an overall score of 0.3571, comprising a relevance score of 0.5365 and a factuality score of 0.1777. As highlighted in Table~\ref{tab:caption_prediction_leaderboard}, this submission secured the third-place ranking on the official leaderboard and successfully attained the highest factuality score of any participating team.

Our concept anchoring approach, described in Section~\ref{sec:concept_anchoring}, reveals a consistent pattern: concept anchoring universally improved the factuality average across all three models, but its impact on overall score is architecture-dependent. InstructBLIP shows a substantial overall gain (+0.028); LLaVA-Med shows a moderate gain (+0.010); BLIP shows essentially no net gain ($-$0.002), because anchoring trades roughly equal amounts of relevance for factuality. By expanding the generation limit to 80 tokens and applying the Vizwins merging algorithm, we achieved a highly competitive BLIP test score of 0.3564.

Our MedGemma-4B zero-shot submission, described in Section~\ref{sec:medgemma-zsp}, was one of the final Task~2 submission.
On the validation split, the PMC-legend prompt with greedy decoding and \texttt{clean\_caption} post-processing reached ROUGE-1\,$=0.2389$, BERTScore F1\,$=0.6205$, and BLEURT\,$\approx 0.44$.
On the held-out test set, the same configuration achieved an overall score of $0.3186$ (Rel.\ $0.5120$, Fact.\ $0.1253$), with BERTScore\,$=0.5908$, ROUGE-1\,$=0.2472$, image--caption similarity $=0.8798$, BLEURT\,$=0.3301$, MedCAT $F_1=0.1281$, and AlignScore\,$=0.1225$.
Despite using a model roughly $7\times$ smaller than the Gemma-3 27B winner and no fine-tuning at all, it lands within $0.04$ of the team's strongest overall score ($0.3571$).

The three MedGemma-4B fine-tuning variants underperformed this zero-shot baseline on the primary caption metrics: Full SFT reached ROUGE-1\,$=0.2119$, BERTScore\,$=0.6079$, BLEURT\,$=-0.05$; Joint Concept\,+\,Caption SFT reached ROUGE-1\,$=0.2038$, BERTScore\,$=0.6112$, BLEURT\,$=0.04$.
The pattern is consistent---fine-tuning preserved or slightly improved BERTScore (semantic content) but tanked BLEURT (style fluency), which we attribute to MedGemma's pretraining already covering the PMC caption distribution well enough that further supervised pressure overwrites style with content.
This is the reason none of the fine-tuned variants were submitted.

Among the post-hoc BiomedCLIP-based candidate pools, raw image-to-text retrieval scored validation ROUGE-1\,$=0.1888$, lifted to $0.2001$ by the ridge-regression alignment; the few-shot RAG pool reached $0.2440$.
Both are below the zero-shot submission standalone, but they catch different images: when we pick the best caption per image across all four pools (zero-shot, RAG, KNN, BiomedCLIP), it raises validation ROUGE-1 from $0.3027$ to $0.3154$.
The raw BiomedCLIP-argmax reference-free selector already lifts validation ROUGE-1 to $0.2449$, and tuning the source biases and length target on a held-out portion of validation pushes it to $0.2510$---about one ROUGE-1 point above the best single system, without any new model training or inference.
This selector was not submitted but is reported as the strongest post-hoc configuration we found.

\section{Discussion}

\paragraph{Task 1: Architectural Diversity and Threshold Generalization.}
Our first-place performance in the Concept Detection task underscores the importance of a holistic pipeline tailored to the specific challenges of medical multi-label classification. First, pruning the label space to the top 1{,}000 most frequent CUIs proved crucial. By intentionally discarding the extreme long tail, we concentrated the ensemble's capacity on concepts with sufficient training support, avoiding the destabilizing gradients and poor calibration associated with rare labels. This stable foundation was further reinforced by Test-Time Augmentation (TTA). The +0.0306 $F_1$ lift over our unaugmented baseline demonstrated that controlled geometric perturbations are highly effective for improving feature invariance in radiology images. Second, the performance gap between our best individual model (ConvNeXt-V2, $F_1$ = 0.5923) and the final three-way ensemble ($F_1$ = 0.6061) highlights the strategic value of architectural diversity. Rather than ensembling homogeneous networks, we fused state-of-the-art Convolutional Neural Networks (ConvNeXt-V2 and DenseNet-169) with a Vision Transformer (BiomedCLIP ViT-B/16). This specific combination allowed the pipeline to capture both fine-grained, localized anatomical textures via the CNNs, and broad, domain-adapted medical priors derived from PubMed Central via BiomedCLIP. The superiority of this late-fusion approach over our dual-stream cross-attention experiment suggests that decoupling the backbones allows for more stable convergence and captures truly complementary error spaces. Finally, the most critical determinant of our test-set generalization was managing the severe overfitting inherent to threshold optimization. Raw coordinate ascent on the full validation set artificially inflated our expected score, masking decision boundaries on lower-frequency concepts. Our novel ``Honest Threshold Tuning'' pipeline, utilizing a deterministic 50/50 split, a hard global fallback for rare concepts, and proportional threshold blending, successfully bridged this generalization gap. By mathematically regularizing the decision thresholds, this strategy transformed our highly capable ensemble into a robust, test-ready system, ultimately securing the top rank (primary $F_1$ = 0.5790) on the official test set.

Finally, our experimental deployment of an end-to-end fine-tuned Gemma-3 27B model for concept extraction yielded a validation $F_1$ of 0.4893. While this outperformed our unaugmented CNN baseline (0.4789), it fell significantly short of our standalone vision encoders and the final ensemble. This performance gap highlights a fundamental limitation of autoregressive generative models in extreme multi-label classification settings. While LLMs excel at sequential reasoning, forcing them into a deterministic, greedy-decoding path to output hundreds of independent, uncalibrated boolean labels (CUIs) removes the mathematical flexibility required to handle long-tail distributions.

Our other Task~1 submission used a much lighter approach: a training-free KNN over frozen BiomedCLIP embeddings with a precision-oriented decoder.
It scored primary $F_1 = 0.5780$ and secondary $F_1 = 0.9599$ on the test set---very close to the fine-tuned ensemble on primary, and very strong on secondary.
We think this means the primary metric is what really benefits from ensembling and threshold tuning, while the secondary 15-CUI track is mostly solved by a strong domain-pretrained encoder on its own.

\paragraph{Task 2: Foundation Model Scaling and Concept Anchoring.}
Our strongest performance in the Caption Prediction task was achieved by scaling up the foundation model, which natively resolved the inherent trade-off between factual accuracy and grammatical fluency. Fine-tuned with QLoRA and evaluated using 3 beams, our 27B parameter Gemma-3 model achieved our highest overall score of 0.3571 (Relevance: 0.5365, Factuality: 0.1777). This configuration obtained a third-place finish in the competition and garnered the highest factuality score of all submissions. The immense parameter count of the Gemma-3 language backbone inherently possesses the clinical reasoning required to integrate complex anatomical concepts while maintaining natural syntactic flow, effectively bridging the factuality-fluency gap without the need for complex multi-stage pipelines.

Regarding our fine-tuning of the Gemma-3 27B model, we observed a notable performance jump (+0.0125 overall score) when shifting from a 1.5-epoch, 2-beam configuration to a 2-epoch, 3-beam configuration. We hypothesize that the expanded beam width played a disproportionate role in the observed factuality improvement (from 0.1644 to 0.1777). Medical captioning is highly sensitive to early token selection. Thus, expanding the hypothesis space to 3 beams likely prevented the auto-regressive decoder from committing to phrasing that were grammatically fluent but clinically inaccurate. Simultaneously, completing the full second epoch of fine-tuning provided the necessary gradient steps to solidify the model's alignment with the specific style of the ROCOv2 captions without overfitting on the training data.

In contrast, our experiments with smaller, resource-constrained models demonstrated that without this scale, bridging that gap requires structured interventions like concept anchoring and output merging. A notable finding from our cross-architecture comparison is that concept anchoring's effectiveness is not uniform across model designs. The encoder-decoder InstructBLIP benefits most (+0.028 overall), likely because structured text is incorporated as encoder input. The LoRA-adapted LLaVA-Med shows moderate gains (+0.010). The fully fine-tuned BLIP shows no net overall gain despite consistent factuality improvements, which we hypothesize is because its decoder-only BERT text head treats the anchor prompt as a description target rather than conditioning context, producing terser captions that are factually correct but less semantically rich. To address this limitation, we applied our custom Vizwins merging algorithm. By dynamically selecting between the semantically rich visual-only captions and the highly factual concept-anchored captions on a per-image basis, we successfully bypassed the architectural constraints of the decoder. This heavily engineered, merged BLIP pipeline achieved a highly competitive overall score (0.3564) that fell only slightly short of the Gemma-3 model, demonstrating that concept anchoring remains a vital strategy when scaling the VLM architecture is not feasible.

MedGemma-4B zero-shot is an interesting experiment at the small end.
With no fine-tuning and just a PMC-legend prompt, it reached an overall test score of $0.3186$---below Gemma-3 27B, but at roughly $7\times$ smaller.
The bigger surprise is that all three fine-tuning variants of the same 4B model (QLoRA attention, QLoRA attention\,+\,MLP, and full SFT) actually hurt BLEURT compared to zero-shot.
Our observation is that MedGemma was already pretrained on PubMed Central, so a domain-matched prompt is enough to hit the right caption style, and additional supervised training on ROCOv2 ends up trading style for content.

\section{Future Work}

Several clear directions emerge from this work.
For Concept Detection (Task~1), the next priority is handling the long tail of medical concepts more carefully.
Instead of dropping rare CUIs from the label space, we plan to explore few-shot learning and hierarchical label grouping, so the model can still learn from rare concepts without the overfitting we saw during threshold tuning.
On the retrieval side, our BiomedCLIP KNN pipeline could be extended with a concept-aware reranker that uses the modality and anatomy probes as priors, which might close the small gap to the three-way ensemble on the primary track.

For Caption Prediction (Task~2), we want to simplify the pipeline rather than add to it.
Two directions in particular: mixed-prompt training, where the model sees both concept-anchored and visual-only inputs during fine-tuning, and a text-only editor pass that merges the most factual and fluent fragments across multiple generated candidates.
We are also interested in seeing whether the next generation of foundation models---the upcoming Gemma-4 series, or successors to MedGemma---natively close the remaining clinical-factuality gap without requiring concept anchoring or merging at all, since scaling to Gemma-3 27B already bypassed our custom caption-merging step.
Extending the reference-free BiomedCLIP reranker to include these stronger candidates is the cheapest way to put this hypothesis on a fair benchmark.

Finally, to support the newly introduced explainability subtask~\cite{damm2025imageclef}, we aim to integrate visual explanation tools, such as attention maps and GradCAM~\cite{selvaraju2017gradcam}, to visually highlight which specific regions of an image drive the model's predictions.

\section{Conclusions}

These working notes detailed our top-performing pipelines for the ImageCLEFmedical Caption 2026 tasks. In Concept Detection, we secured the first-place ranking (primary $F_1$ = 0.5790) by directly targeting the challenges of extreme label imbalance. By coupling a diverse three-way vision ensemble with our regularized Honest Threshold Tuning strategy, we successfully mitigated the long-tail overfitting that typically degrades medical multi-label classification, ensuring robust generalization to unseen test data.

In Caption Prediction, we explored a spectrum of approaches: a concept-anchored cross-architecture ablation across smaller VLMs that evolved into a fully fine-tuned BLIP pipeline with Vizwins merging (0.3564), and a QLoRA fine-tuned Gemma-3 27B model decoded with 3-beam search that reached our strongest overall score of 0.3571.
The contrast suggests that for medical caption generation, foundation-model scale and pretraining-distribution match are the dominant levers, while concept anchoring and output merging remain valuable strategies for smaller, resource-constrained models.

\section*{Acknowledgements}

We thank the Data Science at Georgia Tech (DS@GT) CLEF competition group for their support.
This research was supported in part through research cyberinfrastructure resources and services provided by the Partnership for an Advanced Computing Environment (PACE) at the Georgia Institute of Technology, Atlanta, Georgia, USA~\cite{pace2017}.

\section*{Declaration on Generative AI}
 During the preparation of this work, the author(s) used ChatGPT and Claude Opus 4.7 in order to perform grammar and spelling check.
 After using these tool(s)/service(s), the author(s) reviewed and edited the content as needed and take(s) full responsibility for the publication’s content. 
 
\section*{Code Availability}

To ensure full reproducibility of our results, the source code for both Concept Detection and Caption Prediction tasks in this study is publicly available at: https://github.com/dsgt-arc/imageclef-caption-2026

\bibliography{main}
\end{document}